\documentclass[letterpaper]{article}
\usepackage{spconf,graphicx, amsfonts}
\usepackage{epstopdf}
\usepackage[fleqn]{amsmath}
\usepackage[lined,boxed]{algorithm2e}
 \pdfoutput=1
\usepackage[english]{babel}
\usepackage{fancyhdr}
\usepackage{threeparttable}
\usepackage{parskip}
\usepackage{cite}

\usepackage{caption}
\usepackage{subfig}
\usepackage{float}
\usepackage{graphicx}
\usepackage{tablefootnote}
\usepackage{booktabs}
\usepackage{multirow}
\usepackage[T1]{fontenc}
\usepackage{pgf}
\usepackage[normalem]{ulem}
\usepackage[inline]{enumitem}
\usepackage[autoplay]{animate}

\definecolor{colorsrc}{rgb}{0.36, 0.54, 0.66}

\definecolor{colorwnd2}{rgb}{0.91, 0.84, 0.42}
\definecolor{colorwnd}{rgb}{0.8, 0.0, 0.1}
\definecolor{colorfdd}{rgb}{0.44, 0.16, 0.39}
\definecolor{colorshi}{rgb}{0.55, 0.71, 0.0}
\definecolor{colornan}{rgb}{0.8, .33, 0}
\definecolor{colornan}{rgb}{0.72, 0.53, 0.04}
\definecolor{darkcyan}{rgb}{0.0, 0.55, 0.55}
\definecolor{colorsdd}{rgb}{0.0, 0.2, 1.0}
\definecolor{colorlck}{rgb}{0.0, 0.9, 0.9}
\definecolor{pinegreen}{rgb}{0.0, 0.47, 0.44}

\def\x{{\mathbf x}}

\newcommand{\norm}[1]{\left\|#1\right\|}

\def\bmt{\left[\begin{matrix}}

\def\emt{\end{matrix}\right]}

\def\bx{\mathbf{x}}

\def\bm{\mathbf{m}}

\def\by{\mathbf{y}}
\def\and{\text{~and~}}

\def\etal{\textit{et al.}}
\def\R{\mathbb{R}}

\def\bD{\mathbf{D}}
\def\lbD{\overline{\mathbf{D}}}

\def\bM{\mathbf{M}}

\def\bW{\mathbf{W}}
\def\bX{\mathbf{X}}
\def\lbX{\overline{\mathbf{X}}}
\def\bY{\mathbf{Y}}

\def\rank{\text{rank}}

\def\bW{\mathbf{W}}

\def\bXc{\bX^{0}}

\def\wt{\widetilde}

\def\ben{\begin{equation*}}
\def\een{\end{equation*}}
\def\beaa{\begin{eqnarray*}}
\def\eeaa{\end{eqnarray*}}
\def\bea{\begin{eqnarray}}
\def\eea{\end{eqnarray}}

\makeatletter
\newsavebox\myboxA
\newsavebox\myboxB
\newlength\mylenA

\newcommand*\lbar[2][0.75]{%
    \sbox{\myboxA}{$\m@th#2$}%
    \setbox\myboxB\null
    \ht\myboxB=\ht\myboxA%
    \dp\myboxB=\dp\myboxA%
    \wd\myboxB=#1\wd\myboxA
    \sbox\myboxB{$\m@th\overline{\copy\myboxB}$}
    \setlength\mylenA{\the\wd\myboxA}
    \addtolength\mylenA{-\the\wd\myboxB}%
    \ifdim\wd\myboxB<\wd\myboxA%
       \rlap{\hskip 0.5\mylenA\usebox\myboxB}{\usebox\myboxA}%
    \else
        \hskip -0.5\mylenA\rlap{\usebox\myboxA}{\hskip 0.5\mylenA\usebox\myboxB}%
    \fi}
\makeatother

\def\bbx{\lbar{\bx}}        
\def\bbY{\lbar{\bY}}        
\def\bbD{\lbar{\bD}}        
\title{\vspace{-.2in}LEARNING A LOW-RANK SHARED DICTIONARY \\FOR OBJECT CLASSIFICATION\vspace{-0.1in}}
\name{\vspace{-0.25in}Tiep H. Vu, Vishal Monga \thanks{\noindent \hspace{-0.2in}Research was supported by an Office of Naval Research Grant no. \bf{0401531}.}}

\address{\vspace{-0.1in} {\small Pennsylvania State University, University Park, PA}}

\begin{document}
\maketitle
\thispagestyle{fancy}
\pagestyle{empty}
\begin{abstract}
\vspace{0.1in}
\label{abstract}
Despite the fact that different objects possess distinct class-specific features, they also usually share common patterns. Inspired by this observation, we propose a novel method to explicitly and simultaneously learn a set of common patterns as well as class-specific features for classification. Our dictionary learning framework is hence characterized by both a shared dictionary and particular (class-specific) dictionaries. For the shared dictionary, we enforce a low-rank constraint, i.e. claim that its spanning subspace should have low dimension and the coefficients corresponding to this dictionary should be similar. For the particular dictionaries, we impose on them the well-known constraints stated in the Fisher discrimination dictionary learning (FDDL). Further, we propose a new fast and accurate algorithm to solve the sparse coding problems in the learning step, accelerating its convergence. The said algorithm could also be applied to FDDL and its extensions. Experimental results on widely used image databases establish the advantages of our method over state-of-the-art dictionary learning methods.

\end{abstract}


\vspace{-0.1in}
\section{Introduction}
\vspace{-0.1in}

\begin{figure}[t]
\centering
\includegraphics[width = 0.48\textwidth]{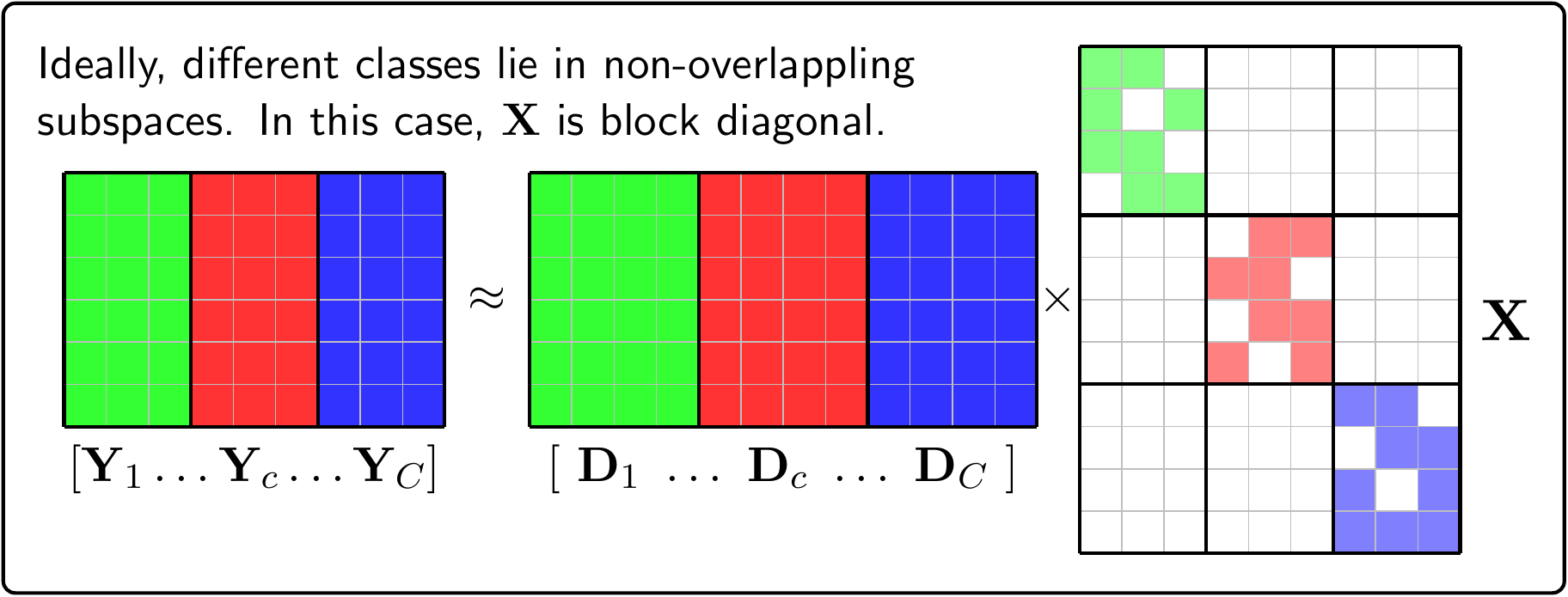}
\caption{\small Ideal structure of the coefficient matrix in SRC.}
\label{fig:srcidea}
\end{figure}

\label{sec:intro}
Sparse representations have emerged as a powerful tool for a range of signal processing applications. Applications include compressed sensing, signal denoising, image inpainting and more recently, signal classification. In such representations, most of signals can be expressed by a linear combination of few bases taken from a ``dictionary''. Based on this theory, a sparse representation classifier\cite{Wright2009SRC} (SRC) was developed for robust face recognition, and was later adapted to several signal/image classification problems\cite{Srinivas2014SHIRC,Mousavi2014ICIP,Srinivas2013,zhang2012multi,thongkamwitoon2015image}. The central idea in SRC is to represent a test sample (e.g. a face) as a linear combination of samples from the available training set. Sparsity manifests because most of non-zeros correspond to bases whose memberships are the same as the test sample. Therefore, in the ideal case, each object is expected to lie in its own class subspace and all class subspaces are non-overlapping.
Concretely, given $C$ classes and a dictionary $\bD = [\bD_1, \dots, \bD_C]$ with $\bD_c$ comprising training samples from class $c, c = 1, \dots, C$, a new sample $\by$ from class $c$ can be represented as $\by \approx \bD_c\bx^c$. Therefore, if we express $\by$ using the dictionary $\bD: \by \approx \bD\bx = \bD_1\bx^1 + \dots + \bD_c\bx^c + \dots + \bD_C\bx^C$, then most of active elements of $\bx$ should be located in $\bx^c$ and hence, the coefficient vector $\x$ is expected to be sparse. In matrix form, let $\bY = [\bY_1, \dots, \bY_c, \dots, \bY_C]$ be the set of all samples where $\bY_c$ comprises those in class $c$, the coefficient matrix $\bX$ would be sparse. In the ideal case, $\bX$ is block diagonal (see Fig. \ref{fig:srcidea}).


\par
It has been shown that learning a dictionary from the training samples instead of using all of them as a dictionary can further enhance the performance of SRC. Most existing classification-oriented dictionary learning methods try to learn {\em discriminative class-specific dictionaries} by either imposing block-diagonal constraints on $\bX$ or encouraging the incoherence between class-specific dictionaries. Discriminative K-SVD\cite{zhang2010discriminative} and Label-consistent K-SVD\cite{Zhuolin2013LCKSVD} learn the discriminative dictionaries by encouraging a projection of sparse codes $\bX$ to be close to a sparse matrix with all non-zeros being one while satisfying a block diagonal structure as in Fig. \ref{fig:srcidea}.  T. Vu \etal \cite{vu2015dfdl,vu2015tmi} with DFDL and M. Yang \etal\cite{Meng2011FDDL} with FDDL apply Fisher-based ideas on dictionaries and sparse coefficients, respectively. Recently, L. Li \etal\cite{li2014learning} with $D^2L^2R^2$ combined the Fisher-based idea and introduced a low-rank constraint on each sub-dictionary. They claim that such a model would reduce the negative effect of noise contained in training samples.

{\bf Closely Related work and Motivation}: The assumption made by most discriminative dictionary learning methods, i.e. non-overlapping subspaces, is unrealistic in practice. Often objects from different classes share some common features, e.g. background in scene classification. This problem has been partially addressed by recent efforts, namely DLSI\cite{ramirez2010classification} and DL-COPAR\cite{kong2012dictionary}. However, DLSI does not explicitly learn shared features since they are still hidden in the sub-dictionaries. DL-COPAR explicitly learns a shared dictionary $\bD_0$ but suffers from the following drawbacks. First, we contend that the subspace spanned by columns of the shared dictionary must have low rank. Otherwise,  class-specific features may also get represented by the shared dictionary. In the worst case, the shared dictionary span may include all classes, greatly diminishing the classification ability. Second, the coefficients (in each column of the sparse coefficient matrix) corresponding to the shared dictionary should be similar. This implies that features are shared between training samples from different classes via the ``shared dictionary''. In this paper, we develop a new low-rank shared dictionary learning framework (LRSDL) which satisfies the aforementioned properties. We show practical merits of enforcing these constraints are significant.
\par

\par {\bf Contributions:} (1) Our framework is a generalized version of the well-known FDDL\cite{Meng2011FDDL} with the additional capability of capturing shared features, resulting in better performance. (2) We propose a fast and accurate algorithm for the sparse coding step in the learning process, resulting in a flexible and practical learning framework. our algorithm can be applied to speed-up FDDL, and related frameworks, such as $D^2L^2R^2$\cite{li2014learning}, DLRD\_SR\cite{ma2012sparse}, and DLCOPAR \cite{kong2012dictionary}.

\vspace{-0.1in}
\section{Discriminative dictionary learning}
\vspace{-0.05in}
\subsection{Notation} 
\begin{figure}[t]
\centering
\includegraphics[width = 0.48\textwidth]{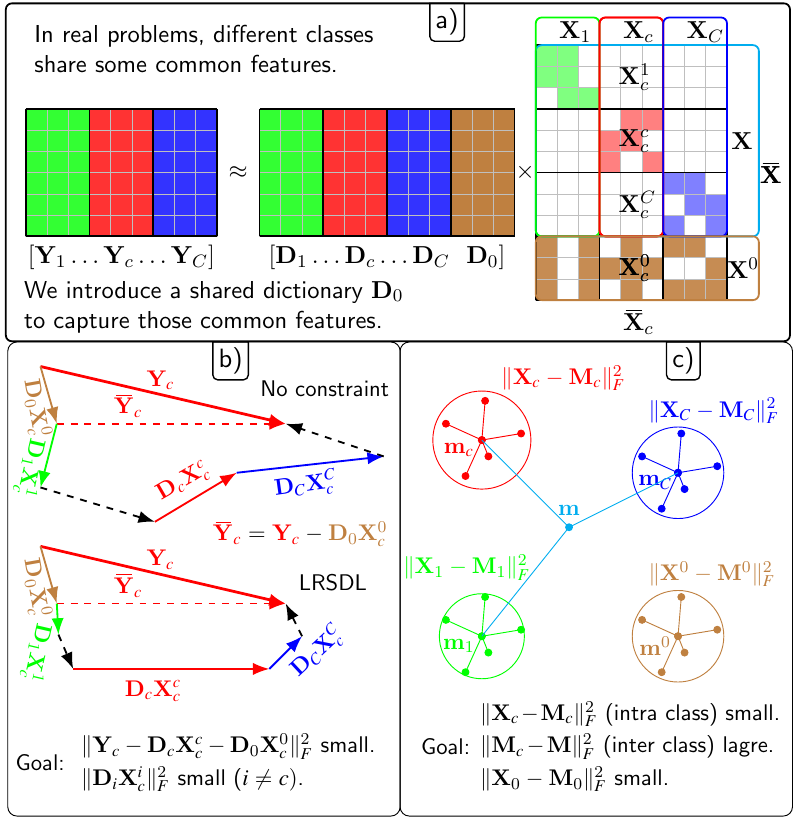}
\caption{\small LRSDL idea with: brown items -- shared; red, green, blue items -- class-specific. a) Notation. b) The discriminative fidelity constraint: class-$c$ sample is mostly represented by $\bD_0$ and $\bD_c$. c) The Fisher-based discriminative coefficient constraint.}
\label{fig:sddlidea}
\end{figure}

\vspace{-0.05in}
\label{sub:notaions}
In addition to notation stated in the Introduction, let $\bD_0$ be the shared dictionary. For $c = 1, \dots, C$; $i = 0, 1, \dots, C$, suppose that $\bY_c \in \R^{d\times n_c}$ and $\bY \in \R^{d\times N}$ with $N = \sum_{c = 1}^C n_c$; $\bD_i \in \R^{d\times k_i}$, $\bD \in \R^{d\times K}$ with $K = \sum_{c=1}^C k_c$; and $\bX \in \R^{K\times N}$. Let $\lbar{\bD} = \bmt\bD & \bD_0\emt$ be the total dictionary.
Denote by $\bX^i$ the sparse coefficient of $\bY$ on $\bD_i$, by $\bX_c \in \R^{{K}\times N_c}$,  the sparse coefficient of $\bY_c$ on ${\bD}$, by $\bX_c^i$ the sparse coefficient of $\bY_c$ on $\bD_i$, $\lbar{\bX} = [\bX^T, (\bX^0)^T]^T$ and $\lbar{\bX}_c = [(\bX_c)^T, (\bX^0_c)^T ]^T$. These ideas are visualized in Fig. \ref{fig:sddlidea}a).
\par
Let $\bm, \bm^0, \bm_c$ be the mean vector of $\bX, \bX^0, \bX_c$, respectively. Let $\bM_c = [\bm_c, \dots, \bm_c] \in \R^{K\times n_c}, \bM^0 =[\bm^0, \dots, \bm^0] \in \R^{k_0\times N}$, and $\bM = [\bm, \dots, \bm]$, with number of columns depending on context, be the mean matrices. The `mean vectors' are illustrated in Fig. \ref{fig:sddlidea}c).
\par Greek letters ($\lambda, \lambda_1, \lambda_2, \eta$) represent positive regularization parameters.
\vspace{-.1in}
\subsection{Fisher discrimination dictionary learning} 
\label{sub:fisher_discrimination_dictionary_learning}
FDDL\cite{Meng2011FDDL} has been used broadly as a technique for exploiting both structured dictionary and learning discriminative coefficient. Particularly, the discriminative dictionary $\bD$ and the sparse coefficient matrix $\bX$ are learned based on minimizing the following cost function:
\vspace{-0.1in}
\begin{eqnarray}
\label{eqn: fddl}
    J(\bD, \bX) = \frac{1}{2}\sum_{c=1}^C r(\bY_c, \bD, \bX_c) + \lambda_1\|\bX\|_1 + \frac{\lambda_2}{2} f(\bX),
\end{eqnarray}
where  $r(\bY_c, \bD, \bX_c) = \|\bY_c - \bD\bX_c\|_F^2 +  \|\bY_c - \bD_c\bX_c^c\|_F^2 + \sum_{i\neq c}\|\bD_i\bX^i_c\|_F^2$, $\sum_{c=1}^Cr(\bY_c, \bD, \bX_c)$ is the discriminative fidelity term, $f(\bX) = \sum_{c=1}^C (\|\bX_c - \bM_c\|_F^2 - \|\bM_c - \bM\|_F^2) + \|\bX\|_F^2$ is the Fisher-based discriminative coefficient term, and the $l_1$-norm encouraging the sparsity of coefficients.
\par The minimization problem in equation (\ref{eqn: fddl}) is solved by alternatively optimizing each $\bX_c$ or $\bD_c$ while fixing all other variables. This approach leads to an extremely slow convergence process which is sometimes impractical for multi-class high dimension problems. Later in this paper, we propose a method to solve all $\{\bX_c\}_{c=1}^{C}$ simultaneously, resulting in a faster algorithm and more accurate solution.

\subsection{Low-rank shared dictionary learning } 
\label{sub:low_rand_shared_dictionary_learning_}
With the presence of the shared dictionary, it is expected that $\bY_c$ can be well represented by the collaboration of the particular dictionary $\bD_c$ and the shared dictionary $\bD_0$. Concretely, the discriminative fidelity term $r(\bY_c, \bD, \bX_c)$ in (\ref{eqn: fddl}) can be extended to $\bar{r}(\bY_c, \lbD, \lbX_c)$ defined as:
\vspace{-0.05in}
\begin{equation*}
    \|\bY_c - \lbar{\bD}\lbar{\bX}_c\|_F^2 + \|\bY_c - \bD_c\bX_c^c - \bD_0\bX^0_c\|_F^2 + \sum_{i = 1, i\neq c}^C \|\bD_i\bX^i_c\|_F^2.
\end{equation*}
Note that $\bar{r}(\bY_c, \lbD, \lbX_c) = r(\lbar{\bY}_c, \bD, \bX_c)$ with $\lbar{\bY}_c = \bY_c - \bD_0\bX^0_c$ (see Fig. \ref{fig:sddlidea}b)).
\par
The Fisher-based discriminative coefficient term $f(\bX)$ is extended to $\bar{f}(\lbar{\bX})$ defined as:
\vspace{-0.05in}
\begin{equation}
\label{eqn:deffbar}
    \lbar{f}(\lbar{\bX}) = f(\bX) + \|\bX^0 -\bM^0\|_F^2,
\end{equation}
where the term $\|\bX^0 -\bM^0\|_F^2$ forces the coefficients of all training samples represented via the shared dictionary to be similar (see Fig. \ref{fig:sddlidea}c)).
\par For the shared dictionary, as stated in the Introduction, we constrain $\rank(\bD_0)$ to be small by using the nuclear norm $\|\bD_0\|_*$ which is its convex relaxation \cite{recht2010guaranteed}. Finally, the cost function $\lbar{J}(\lbar{\bD}, \lbar{\bX})$ of our proposed LRSDL is:
\vspace{-0.1in}
\begin{equation}
\label{eqn:mainsddl}
   \frac{1}{2}\sum_{c=1}^C \lbar{r}(\bY_c, \lbar{\bD}, \lbar{\bX}_c) + \lambda_1\|\lbar{\bX}\|_1 + \frac{\lambda_2}{2} \lbar{f}(\lbar{\bX}) + \eta \|\bD^0\|_*.
\end{equation}
By minimizing this objective function, we can jointly find the appropriate dictionaries as we desire. Notice that if $k_0 = 0$, then $\lbar{\bD}, \lbar{\bX}$ become $\bD, \bX$, respectively, $\lbar{J}(\lbar{\bD}, \lbar{\bX}) $ becomes $J(\bD, \bX)$ and our LRSDL reduces to FDDL.
\par
\noindent \textbf{Classification scheme:} After the learning process, we obtain $\lbar{\bD}, \bm_c, \bm^0$. For a new test sample $\by$, first we find its coefficient vector $\lbar{\bx} = [\bx^T, (\bx^0)^T]^T$ with the sparsity constraint on $\lbar{\bx}$ and further encourage $\bx^0$ to be close to $\bm^0$:
\begin{equation}
\label{eqn:findcodebarx}
    \lbar{\bx} = \arg\min_{\lbar{\bx}} \frac{1}{2}\|\by - \lbar{\bD}\lbar{\bx}\|_2^2 + \frac{\lambda_2}{2}\|\bx^0 - \bm^0\|_2^2 + \lambda_1\|\lbar{\bx}\|_1.
\end{equation}
After finding $\lbar{\bx}$, we extract the contribution of the shared dictionary to obtain $\lbar{\by} = \by - \bD_0\bx^0$. The identity of $\by$ is determined by:
\begin{equation}
    \arg\min_{1 \leq c \leq C} (w\|\lbar{\by} - \bD_c\bx^c\|_2^2 + (1-w)\|\bx - \bm_c\|_2^2),
\end{equation}
where $w \in [0,1]$ is a preset weight for balancing the contribution of the two terms.

\section{Solving optimization problems} 
\label{sub:solving_the_opt}
\par Although the objective function in (\ref{eqn:mainsddl}) is not jointly convex in $\lbar{\bD}, \lbar{\bX}$, it is separably convex with respect to each of $\bX, \bXc, \bD_i$. Therefore, an algorithm that alternatively optimizes each variable while fixing others can be designed. 
\par The particular dictionary {\bf $\bD_c (c = 1, \dots, C)$ update problems} are similar to those in FDDL\cite{Meng2011FDDL} using Online dictionary learning\cite{mairal2010online}, while {\bf the $\bD_0$ update problem} can be written as:
\begin{equation}
    \bD_0 = \arg\min_{\bD_0}  \norm{\frac{\lbar{\bY} + \wt{\bY}}{2} - \bD_0\bX^0}_F^2 + \eta\|\bD_0\|_*,
\end{equation}
which is effectively solved by ADMM\cite{boyd2011distributed} method and the singular value threshoding algorithm\cite{cai2010singular}. We have defined:
\begin{equation}
\label{eqn:defybarytilde}
 \lbar{\bY} = \bY - \bD\bX,~\wt{\bY}_c = \bY_c - \bD_c \bX_c^c,~\wt{\bY} = [\wt{\bY}_1, \dots, \wt{\bY}_C].
\end{equation}

\par The problems of solving $\bX, \bX^0$ in (\ref{eqn:mainsddl}) and $\lbar{\bx}$ in (\ref{eqn:findcodebarx}) can be written in the form:
\begin{equation}
\label{eqn:fista}
    \bW = \arg\min_{\bW} g(\bW) + \lambda\|\bW\|_1,
\end{equation}
where $g(\bW)$ is convex, continuously differentiable with Lipschitz continuous gradient. This family of problems has been shown to be solved effectively using FISTA\cite{beck2009fast}, which is an iterative method that requires calculating the gradient of $g(\bW)$ at each iteration. Next, we present methods to solve for $\bX$, $\bX^{0}$ and $\lbar{\bx}$, where $g(\cdot$) is different in each case and includes terms as in \eqref{eqn:mainsddl} and \eqref{eqn:findcodebarx} excluding the $l_1$ norm term.




{\bf For updating} $\bX$: First we rewrite:
\begin{equation*}
    \sum_{c=1}^C \lbar{r}(\bY_c, \lbar{\bD}, \lbar{\bX}_c)  = \sum_{c=1}^C r(\lbar{\bY}_c, \bD, \bX_c) = \\
\end{equation*}
\vspace{-0.15in}
\begin{equation*}
    \norm{\smash{\underbrace{\bmt
        \lbar{\bY}_1 & \lbar{\bY}_2  & \dots & \lbar{\bY}_C \\
        \lbar{\bY}_1 & \mathbf{0} & \dots & \mathbf{0} \\
        \mathbf{0} & \lbar{\bY}_2 & \dots & \mathbf{0} \\
        \dots & \dots & \dots & \dots \\
        \mathbf{0} & \mathbf{0} & \dots & \lbar{\bY}_C
        \emt}_{\widehat{\bY}}}
   - \underbrace{\bmt
     \bD_1 & \bD_2  & \dots & \bD_C \\
     \bD_1 & \mathbf{0} & \dots & \mathbf{0} \\
     \mathbf{0} & \bD_2 & \dots & \mathbf{0} \\
     \dots & \dots & \dots & \dots \\
     \mathbf{0} & \mathbf{0} & \dots & \bD_C
     \emt}_{\widehat{\bD}}
  \bX
  }_F^2
\end{equation*}
\vspace{-0.1in}
\begin{equation}
  = \| \widehat{{\bY}}  - \widehat{\bD} \bX \|_F^2,
\end{equation}
then we obtain:
  $\displaystyle \frac{\partial \frac{1}{2}\sum_{c=1}^C\lbar{r}(\bY_c, \lbar{\bD}, \lbar{\bX}_c)}{\partial \bX} = \widehat{\bD}^T\widehat{\bD}\bX + \widehat{\bD}^T\widehat{\bY} $. Significantly, although the number of rows of $\widehat{\bY}$ and $\widehat{\bD}$ are extremely high $\left((C+1)\times d\right)$, their number of columns equal those of $\bY$ and $\bD$. Therefore, $\widehat{\bD}^T\widehat{\bD}$ and $\widehat{\bD}^T\widehat{\bY}$ have same dimensions as $\bD^T\bD$ and $\bD^T\bY$, respectively. Moreover, the computational costs of these two terms are low since both $\widehat{\bY}$ and $\widehat{\bD}$ are block sparse.
\par
For $\lbar{f}(\lbar{\bX})$, in case $n_c$'s are all the same\footnote{The same number of training samples in every class is typical in dictionary learning problems. In case $n_c$'s are different, the results change slightly and will be discussed in our future work.}, by using (\ref{eqn:deffbar}) and the result in DFDL\cite{vu2015tmi}, we have:
\begin{equation}
    \frac{\partial \lbar{f}}{\partial \bX} = \frac{\partial f}{\partial \bX} = 4\bX + 2\bM - 4\bmt \bM_1 &\dots & \bM_C\emt,
\end{equation}
which require small computation. The gradient of $g$ in this problem is now inexpensively calculated. The strategy here can be applied to the sparse coding update step in FDDL and its modifications.
\par
{\bf For updating} $\bX^0$, with some mathematical simplifications, we obtain:
\vspace{-0.1in}
\begin{equation*}
    \lbar{J}(\bX^0) = \norm{\frac{\lbar{\bY} + \wt{\bY}}{2} - \bD_0\bX^0}_F^2 + \frac{\lambda_2}{2}\|\bX^0 - \bM^0\|_F^2 + \lambda_1\|\bX^0\|,
\end{equation*}
where $\lbar{\bY}, \wt{\bY}$ are defined in (\ref{eqn:defybarytilde}).
 This optimization problem also has form (\ref{eqn:fista}) where gradient of $g$ is calculated as:
\begin{equation}
    2\bD_0^T\bD_0\bX^0 - \bD_0^T(\bbY + \wt{\bY}) + \lambda_2( \bX^0 - \bM^0).
\end{equation}

\par The final optimization problem is to {\bf solve $\lbar{\bx}$ in  problem } (\ref{eqn:findcodebarx}) with gradient of $g$: $$\bbD^T\bbD\bbx - \bbD\by + \lambda_2\bmt \mathbf{0}\\\bx^0 - \bm^0\emt.$$


\begin{figure}[t]
\centering
\includegraphics[width = 0.4\textwidth]{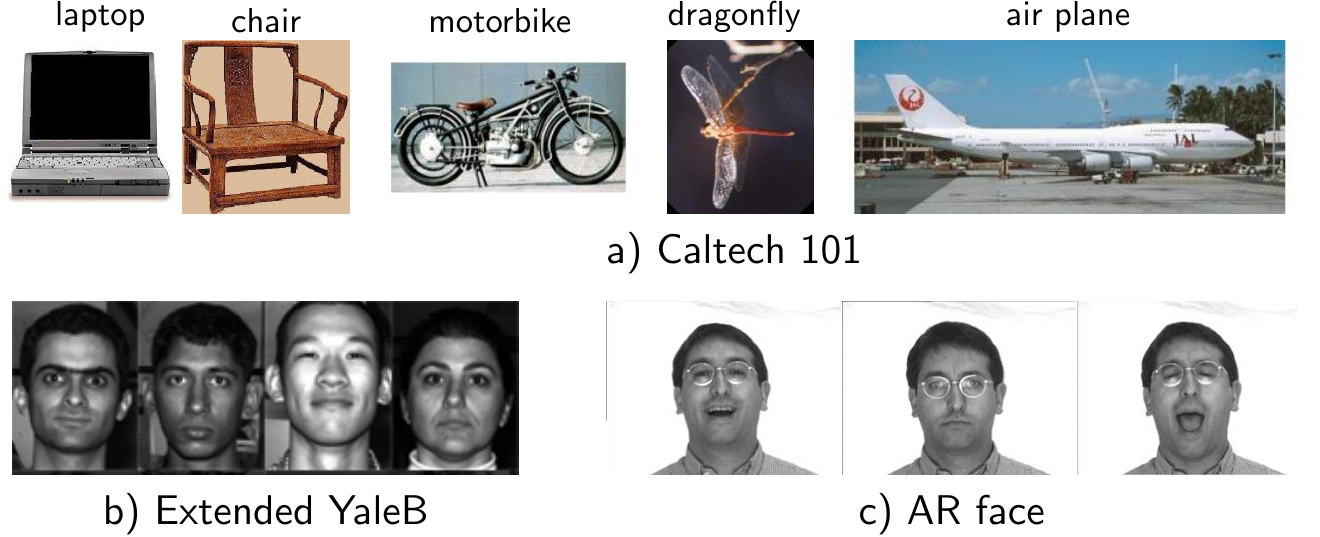}
\caption{\small Examples from three databases. }
\vspace{-0.175in}
\label{fig:examples}
\end{figure}


\vspace{-0.175in}
\begin{figure}[t]
\centering
\includegraphics[width = 0.48\textwidth]{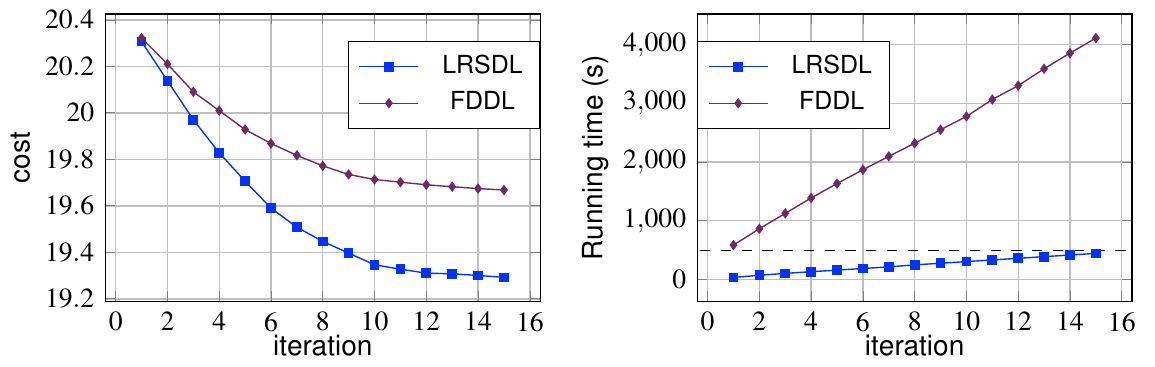}
\caption{\small LRSDL and FDDL convergence rate comparison. }
\label{fig:lrsdlfddlcompare}
\end{figure}

\section{Experimental results} 
\vspace{-0.1in}
\label{sec:experiment_results}

We present the experimental results of applying LRSDL to three diverse databases: the Extended YaleB face database\cite{georghiades2001few}, the AR face database\cite{ardataset}, and one multi-class object category database -- the Caltech 101\cite{fei2007learning}. Example images from these databases are shown in Fig. \ref{fig:examples}.
We compare our results with those using SRC\cite{qi2012robust}, LLC\cite{song2015locality}, and other state-of-the-art dictionary learning methods: LC-KSVD\cite{Zhuolin2013LCKSVD}, DLSI\cite{ramirez2010classification}, FDDL\cite{Meng2011FDDL} and DLCOPAR\cite{kong2012dictionary}.
\par
For two face databases, feature descriptors are random faces, which are made by projecting a face image onto a random vector using a random projection matrix. As in \cite{zhang2010discriminative}, the dimension of a random-face feature in  the Extended YaleB is $d= 504$, while the dimension in AR face is $d = 540$.
\par For the Caltech 101 database, we use a dense SIFT (DSIFT) descriptor. The DSIFT descriptor is extracted from $25\times 25$ patch which is densely sampled on a dense grid with 8 pixels. We then extract the sparse coding spatial pyramid matching (ScSPM) feature\cite{yang2009linear}, which is the concatenation of vectors pooled from words of the extracted DSIFT descriptor. Dimension of words is 1024 and max pooling technique is used with pooling grid of $1\times 1, 2 \times 2$, and $4 \times 4$. With this setup, the dimension of ScSPM feature is 21504; this is followed by dimension reduction to $d = 3000$ using PCA. In experiments, for two shared dictionary learning methods (DLCOPAR\cite{kong2012dictionary} and LRSDL), dictionary sizes are 1120 ($k_c=10,k_0 = 100$) for $n_c = 15$ and 2090 ($k_c = 20, k_0 = 50$) for $n_c = 30$; for other dictionary learning methods, size of the total dictionary is 1530 ($k_c = 15)$ for $n_c = 15$ and 2550 ($k_c = 25$) for $n_c = 30$.


\begin{table}[]
\centering
\small
\caption{Results (\%) on widely used face databases.}
\label{tab:face}
\begin{tabular}{|l||c|c|c|c|}
\hline
Data base       & \multicolumn{2}{c|}{Extended YaleB} & \multicolumn{2}{c|}{AR}        \\ \hline
Training images                     & 15             & 30             & 15             & 20            \\ \hline
\hline
SRC\cite{Wright2009SRC}             & 92.84          & 95.13          & 95.18          & 96.83         \\ \hline
LC-KSVD1\cite{Zhuolin2013LCKSVD}    & 94.50          & 95.60          & 94.18          & 97.8          \\ \hline
LC-KSVD2\cite{Zhuolin2013LCKSVD}    & 95.00          & 96.00          & 94.45          & 97.70         \\ \hline
FDDL\cite{Meng2011FDDL}             & 94.87          & 97.52          & 94.81          & 97.00         \\ \hline
DLSI\cite{ramirez2010classification}& 90.88          & 96.50          & 90.45          & 96.67         \\ \hline
DLCOPAR\cite{kong2012dictionary}    & 94.57          & \textbf{98.03} & 96.81          & 98.5          \\ \hline
\hline
LRSDL                           & \textbf{95.25} & 98.00          & \textbf{96.90} & \textbf{98.7} \\ \hline
\end{tabular}
\end{table}

\begin{table}[]
\centering
\small
\caption{Results (\%) on the Caltech 101 database.}
\label{tab:caltech101}
\begin{tabular}{|l||c|c||c|c|}
\hline
Training images & \multicolumn{2}{c||}{15}                                                                      & \multicolumn{2}{c|}{30 }                                                                      \\ \hline
                & \begin{tabular}[c]{@{}c@{}}Acc. \\ (\%)\end{tabular} & \begin{tabular}[c]{@{}c@{}}Dict. \\ size\end{tabular} & \begin{tabular}[c]{@{}c@{}}Acc. \\  (\%)\end{tabular} & \begin{tabular}[c]{@{}c@{}}Dict. \\ size\end{tabular} \\ \hline
         \hline
SRC\cite{Wright2009SRC}             & 64.26             & 1530         & 73.12              & 3060              \\ \hline
LC-KSVD1\cite{Zhuolin2013LCKSVD}    & 66.70             & 1530         & 73.40              & 2550              \\ \hline
LC-KSVD2\cite{Zhuolin2013LCKSVD}    & 67.70             & 1530         & 73.60              & 2550              \\ \hline
FDDL\cite{Meng2011FDDL}             & 65.56             & 1530         & 73.64              & 2550              \\ \hline
LLC\cite{wang2010locality}          & 65.43             & -            & 73.44              & -                 \\ \hline
DLSI\cite{ramirez2010classification}& 61.28             & 1530         & 70.72              & 2550              \\ \hline
DLCOPAR\cite{kong2012dictionary}    & 67.98             & 1120         & 75.27              & 2090              \\ \hline
\hline
LRSDL                           & \textbf{68.76}    & 1120         & \textbf{76.50}     & 2090              \\ \hline
\end{tabular}
\end{table}
\vspace{-0.15in}
\subsection{LRSDL and FDDL convergence rate comparison} 
\vspace{-0.1in}
\label{sub:compare_fddl_and_lrsdl_runnng}
Before comparing classification accuracy of different methods on different databases, we conduct a toy example on LRSDL and FDDL to a reduced set of training samples from the AR face database to verify the convergence speed of our algorithm. In this example, number of classes $C = 100$, the random-face feature dimension $d = 300$, number of training samples per class $n_c = n = 7$, number of atoms in each particular dictionary $k_c = 7$ (we set $k_0 = 0$ to have exactly the same optimization problems in LRSDL and FDDL); $\lambda_1 = 0.001, \lambda_2 = 0.01$, and number of iterations is 15. Fig \ref{fig:lrsdlfddlcompare} illustrates cost functions and accumulated running time of two methods after each iteration. It is evident that the cost function of our LRSDL is smaller and the gap between two cost functions increases over time. The same trends can be observed from running time of two algorithms. It is significant that total running time of LRSDL after 15 iterations is even smaller than time to run the first iteration in FDDL, thanks to our proposed algorithm presented in Section \ref{sub:solving_the_opt}. With lower cost function, LRSDL provides better overall classification accuracy at 94.13\%, the number in FDDL method is 93.56\%.

\vspace{-0.15in}
\subsection{Classification accuracy comparison} 
\label{sub:exp}
Table \ref{tab:face} shows overall classification results of various methods on two face databases. It is evident that two dictionary learning with shared features (DLCOPAR\cite{kong2012dictionary} and our proposed LRSDL) outperform others by about 0.5\% with three out of four highest values presenting in our proposed LRSDL. For the Caltech 101 database, the same trend is shown in Table \ref{tab:caltech101} with DLCOPAR\cite{kong2012dictionary} and LRSDL outperforming others, by about 2\%, albeit their dictionary sizes are smaller than others, and also, highest classification accuracy is achieved by the proposed LRSDL method.

\newpage
\pagebreak
\bibliographystyle{IEEEbib}
{\small \bibliography{ICIP_2016_Tiep}}
\end{document}